\documentclass{article}
\usepackage{spconf,amsmath,graphicx}

\usepackage{helvet} 
\usepackage{courier}  
\usepackage{graphicx} 
\usepackage{caption} 

\usepackage{graphicx}
\usepackage{booktabs}
\usepackage{subfigure}
\usepackage{algorithm}
\usepackage{algorithmic}
\usepackage{color}

\usepackage{microtype}
\usepackage{amsmath, amssymb, amscd, amsfonts}
\usepackage{IEEEtrantools}
\usepackage{soul}
\usepackage{url}
\usepackage{amsmath}
\usepackage{algorithm}
\usepackage{algorithmic}
\usepackage{amssymb}
\usepackage{amsfonts} 
\usepackage{multirow}
\usepackage{boldline}
\usepackage{hhline}
\usepackage{xcolor}

\usepackage{makecell}
\usepackage{mathrsfs}
\usepackage[normalem]{ulem}
\useunder{\uline}{\ul}{}
\usepackage[inline]{enumitem}
\usepackage{enumitem}
\usepackage{bbding}
\usepackage{pifont}

\title{A Soft Contrastive Learning-based Prompt Model for Few-shot Sentiment Analysis}
\name{Jingyi Zhou$^\dagger$, Jie Zhou$^{\natural\star}$\thanks{$^\star$ Corresponding authors, jzhou@cs.ecnu.edu.cn.}, Jiabao Zhao$^\mathsection$, Siyin Wang$^\dagger$, Haijun Shan$^\ddag$, Gui Tao$^\dagger$, Qi Zhang$^{\dagger\star}$, Xuanjing Huang$^\dagger$}
\address{$^\dagger$School of Computer Science, Fudan University; $^\ddag$CEC GienTech Technology Co.,Ltd.; 
\\ $^\natural$
School of Computer Science and Technology, East China Normal University, Shanghai, China; 
\\ $^\mathsection$
Lab of Artificial Intelligence for Education, East China Normal University, Shanghai, China 
}
\begin{document}
%
\maketitle
\begin{abstract}
Few-shot text classification has attracted great interest in both academia and industry due to the lack of labeled data in many fields. 
Different from general text classification (e.g., topic classification), few-shot sentiment classification is more challenging because the semantic distances among the classes are more subtle. 
For instance, the semantic distances between the sentiment labels in a positive or negative polarity (e.g., ``love" and ``joy", ``remorse" and ``sadness") are close, while the distances are large for the sentiment labels in two opposite polarities (e.g., ``love" and ``sadness"). 
To address this problem, we propose a Soft Contrastive learning-based Prompt (\texttt{SCP}) model for few-shot sentiment analysis.
First, we design a sentiment-aware chain of thought prompt module to guide the model to predict the sentiment from coarse grain to fine grain via a series of intermediate reasoning steps.
Then, we propose a soft contrastive learning algorithm to take the correlation of the labels into account.
A series of experiments on several sentiment analysis datasets show the great advantages of \texttt{SCP} by comparing it with SOTA baselines (e.g., ChatGPT).
\end{abstract}
\vspace{-2mm}
\section{Introduction}
\label{sec:intro}
\vspace{-1mm}
Due to the surge of online reviews and the advancement of deep learning techniques, sentiment classification has achieved great success in the past decade \cite{liu2012sentiment}. 
Most of the existing methods heavily rely on large-scale labeled data for each class, which is costly and time-consuming. 
In this paper, we focus on few-shot sentiment classification, which predicts classes with few labeled data. Instead of increasing the amount of data or stacking model structures, we design a contrastive learning method with a designed training strategy adapted to sentiment classification tasks, utilizing limited data to fully stimulate the potential of the language model.

\begin{figure}[!t]
\centering
\includegraphics[width=6.5cm,height=6cm]{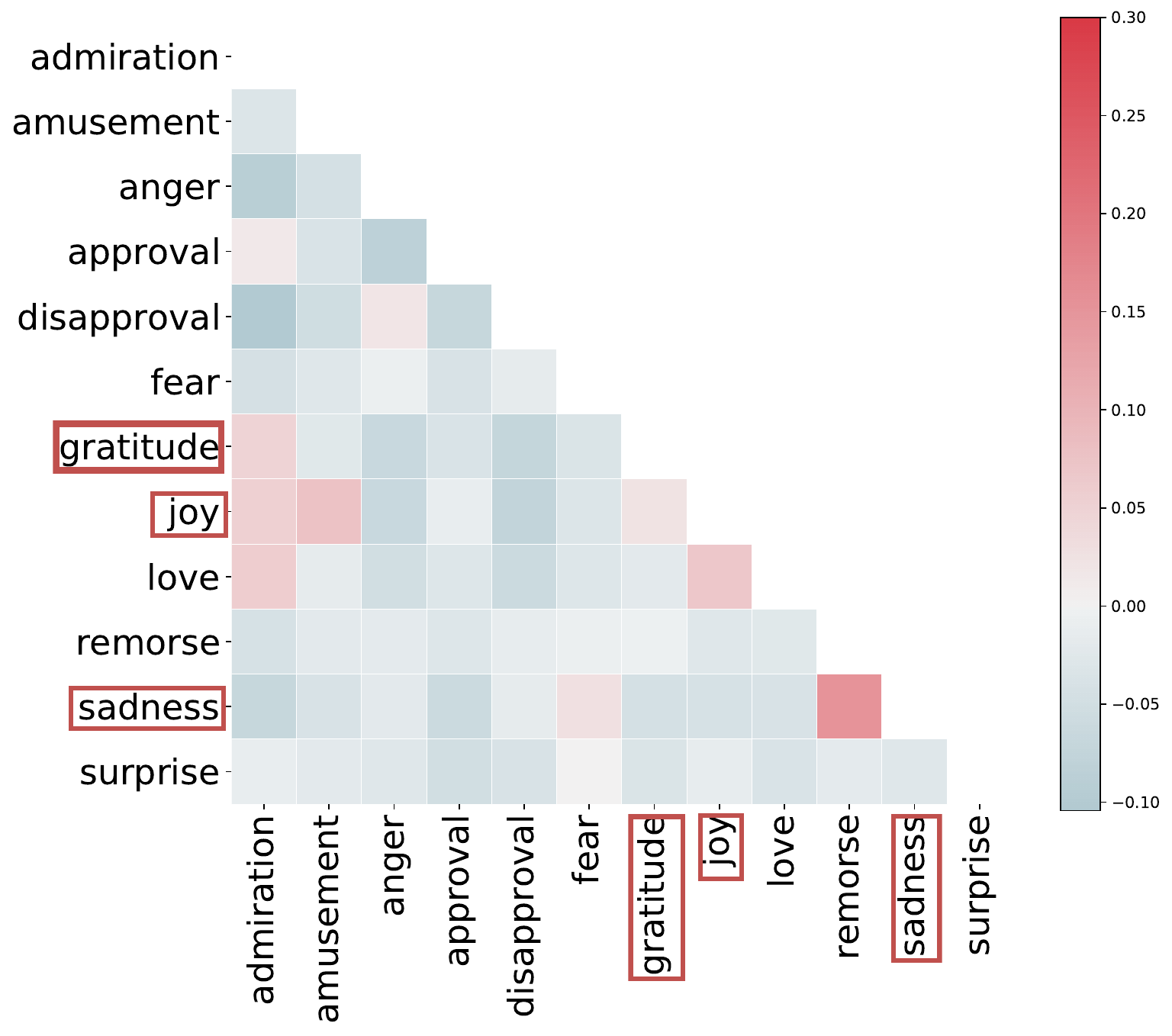}
\vspace{-2mm}
\caption{The correlation heatmap. Red and blue mean the positive and negative correlation respectively. Color depth indicates the degree of correlation.}
\label{fig:correlation}
\vspace{-5mm}
\end{figure}

The existing methods often regard this task as a few-shot text classification task, such as data augmentation-based method \cite{chen2020mixtext,xie2020unsupervised} and prompt-based method \cite{Schick2021ExploitingCF,hu2021knowledgeable}. However, these studies ignore the critical characteristics of fine-grained sentiment classes in sentiment classification, leading to limited performances. 
Compared to text classification tasks (e.g., topic classification), the sentiment semantics expressed by the users may be similar and fine-grained (e.g., joy, love), which makes the task more challenging. 
Inspired by the prior knowledge provided by \cite{Demszky2020GoEmotionsAD}, we calculate the Pearson correlation values between each pair of emotions using average judgment vectors of raters. As shown in Figure \ref{fig:correlation}, the semantic distances between sentiment classes vary a lot. The semantic distances between the sentiment labels in a positive or negative polarity (e.g., ``love" and ``joy", ``remorse" and ``sadness") are close, while the distances between the sentiment labels in two opposite polarities (e.g., ``love" and ``sadness") are large.
Hence, the correlations among the emotion classes are important for few-shot sentiment classification.

\begin{figure*}[t!]
\centering
\includegraphics[scale=0.27]{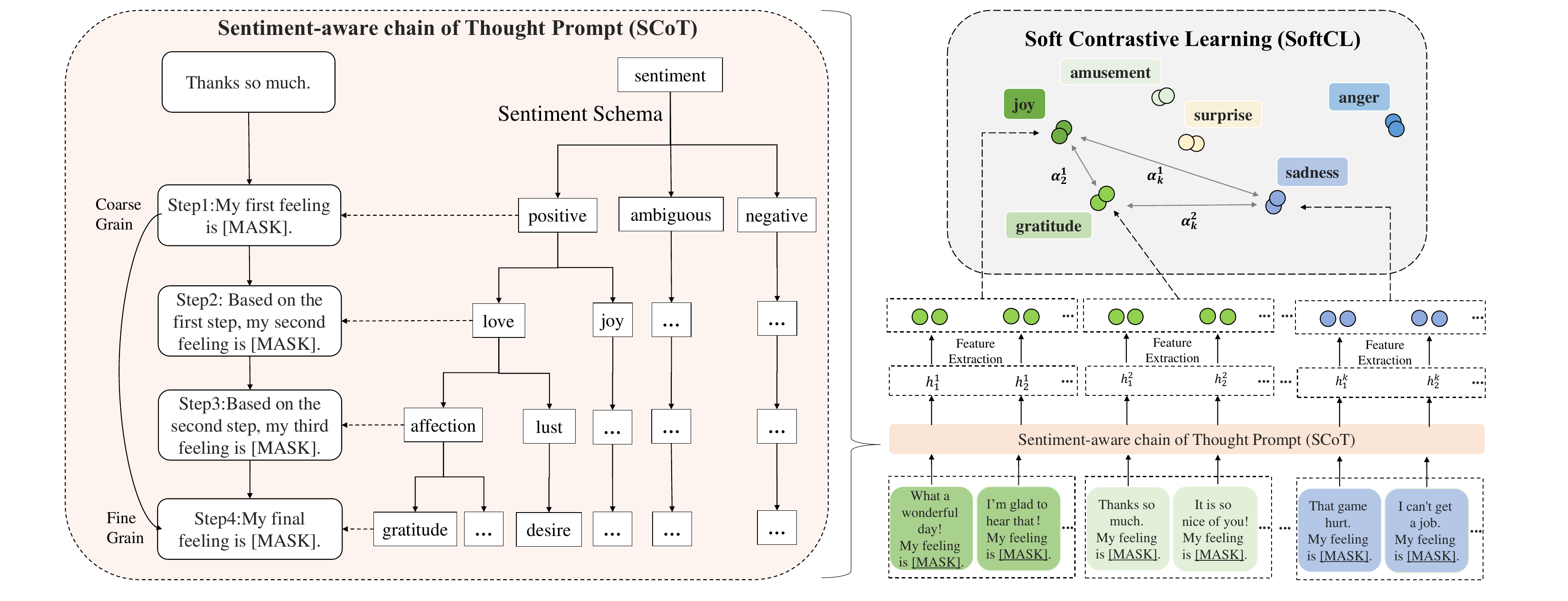}
\vspace{-1mm}
\caption{The architecture of our proposed \texttt{SCP} model for few-shot sentiment analysis. 
}
\label{fig:framework}
\vspace{-3mm}
\end{figure*}

To better take advantage of the characteristics of sentiment classification and exploit the capability of language models, incorporating contrastive learning into a prompt-tuning framework.
First, inspired by the process of human thought \cite{DBLP:conf/iclr/ZhouSHWS0SCBLC23}, we design a sentiment-aware chain of thought prompt strategy to infer the sentiment from coarse grain to fine grain.
Then, to consider the correlation of sentiment labels, we introduce a soft contrastive learning method to model these fine-grained sentiments.
Traditional supervised contrastive learning is not compatible for sentiment analysis because it simply regards the samples of other classes as the negative samples hardly \cite{khosla2020supervised}.
However, the relations between samples with different classes are more complex while two samples with different classes (e.g., joy, happy) may be emotionally close to each other.
Thus, we integrate the relationships among the labels into contrastive learning in a soft way to consider the distances between two samples, which helps the implicit learning of correlations among sentiments.

The main contributions are summarized as follows: 
1) We propose a \texttt{SCP} model for few-shot sentiment classification, which integrates contrastive learning into a sentiment-aware chain of thought prompt framework; 
2) We design a soft contrastive learning mechanism by considering the distances of positive and negative samples softly; 
3) Extensive experiments are conducted, showing our model can effectively improve the performance of few-shot sentiment analysis.

\vspace{-2mm}
\section{Our Approach}
\label{sect:our approach}
\vspace{-1mm}
In this paper, we propose a \texttt{SCP} model for few-shot sentiment analysis (Figure \ref{fig:framework}).
To stimulate the model's ability of sentiment reasoning, we design a sentiment-aware chain of thought prompt tuning architecture with a soft contrastive learning mode. 
Specifically, we design a sentiment-aware chain of thought prompt model to classify fine-grained sentiment step by step using the guided prompts. Additionally, to consider the distances between the positive and negative samples belonging to different classes softly, we propose a soft contrastive learning method via the correlation of sentiment labels.

Formally, given a corpus $\mathcal{D}=\{(x_i, y_i)\}_{i=1}^{|\mathcal{D}|}$, where $|\mathcal{D}|$ is the number of samples in $\mathcal{D}$.
$x$ is a sentence with words $\{w_{1},...,w_{|x|}\}$ and $y \in C$ is $x$'s sentiment label. 
Here, $C$ is the set of sentiment classes where each class is a sentiment word. 

\vspace{-1mm}
\subsection{Sentiment-Aware Chain of Thought Prompt}
To better leverage the knowledge in pre-trained language models (e.g., BERT \cite{devlin-etal-2019-bert}), we introduce a prompt-based model, which has obtained great attention for few-shot tasks in the field of NLP \cite{Schick2021ExploitingCF,hu2021knowledgeable,Wei2022ChainOT}. 
Particularly, we propose a sentiment-aware chain of thought (SCoT) prompt method for few-shot sentiment analysis tasks inspired by Least-to-Most proposed by \cite{DBLP:conf/iclr/ZhouSHWS0SCBLC23}.
This module generates a series of intermediate reasoning steps via a sentiment schema containing the detailed definition of basic emotion, secondary emotion, and tertiary emotion \cite{parrott2001emotions}, which classifies emotions into multiple levels from coarse grain to fine grain. It allows the model to decompose sentiment prediction into several intermediate steps with more reasoning steps, which is close to the human way of thinking about sentiments. 

Specifically, we judge the sentiment of a sentence $x$ into four steps. For the $t$-th step, we combine the former prompt sentence $\mathcal{T}_{t-1}(x)$ with the $t$-th template $\mathcal{T}_t$ to get $\mathcal{T}_t(x)$ that contains one mask token. In particular, $\mathcal{T}_0(x)=x$.

For example, if ${T}_0(x)$ = ``The food is so delicious!", then $\mathcal{T}_{1}(x)$= ``${T}_0(x)$ My first feeling is [MASK]."; $\mathcal{T}_{2}(x)$= ``${T}_1(x)$ Based on the first step, my second feeling is [MASK]."; $\mathcal{T}_{3}(x)$= ``${T}_2(x)$ Based on the second step, my third feeling is [MASK]."; $\mathcal{T}_{4}(x)$= ``${T}_3(x)$ My final feeling is [MASK].". \footnote{The templates we used are simple since they perform well, and we would like to explore more complex templates in future work.} 

The same as that in the pre-training phase, we finetune the pre-trained model $\mathcal{M}$ by predicting the masked token. 
Specifically, we input the prompt into $\mathcal{M}$ and use the masked language model \cite{devlin-etal-2019-bert} to predict the sentiment based on the representation of the masked token. 
\vspace{-2mm}
\begin{equation}
    p^{\mathrm{mask}}_t(x) = \mathrm{MLM}(\mathcal{M}(\mathcal{T}_t(x)))
\end{equation}
where MLM is a masked language model that is a simple linear layer with softmax, $\mathcal{M}(\cdot)$ is used to calculate the representation of the masked token.

We use cross-entropy loss between the predicted probability distribution $p^{\mathrm{mask}}_t(x)$ and the ground truth $y$,
\vspace{-2mm}
\begin{equation}
    \mathcal{L}_{p,i} = - \sum_{t}\log(p^{\mathrm{mask}}_t(y_i|x_i))
\end{equation}
where $p^{\mathrm{mask}}_t(y_i|x_i)$ is the probability of class $y_i$. 
The process tries to imitate the thinking process of human recognition, enabling the model to have a deeper recognition of sentiments.  


\vspace{-1mm}
\subsection{Soft Contrastive Learning}
Supervised contractive learning \cite{khosla2020supervised} aims to pull the points belonging to the same class together and simultaneously push apart clusters of samples from different classes hardly. However, it ignores the nuanced relations between classes, limiting the model's deeper understanding of sentiment classes. 
Fortunately, the correlation of the sentiment labels can describe the distances between the positive samples and negative samples belonging to different classes. 
Thus, we propose a soft contrastive learning (SoftCL) method that takes the correlation of the sentiment labels into account. 

Let $h \in \mathbb{R}^{|C|}$ be the representation of the sentence $x$, which is intercepted from the output vector of [cls], where $|C|$ is the length of the label set.
Consider a sequence of samples $X=\left\{x_1 ,...,x_n\right\}$, $X$ is arranged in the order of input. 
Let $H=\left\{h_1,...,h_n\right\}$ be the representation sequence of $X$, and $Y=\left\{y_1,...,y_n\right\}$ be the labels of $X$. $\forall x_i \in X$,  $h_i \in H$ is the representation of $x_i$, $y_i \in Y$ is the label of $x_i$. 
We calculate contrastive loss between each sample pair $(x_i,x_j),j<i$. With the distance between representation vectors measured by cosine distance, a correlation-weighted contrastive loss function is proposed:
\begin{equation}
\vspace{-1mm}
\nonumber
\small
\begin{aligned}
        \mathcal{L}_{\mathrm{SoftCL},i} & = \dfrac{-1}{|S(i)|} \log \sum_{\substack{k \in S(i)}} \dfrac{\dfrac{1}{\alpha_{y_k}^{y_i}} exp(h_i \cdot h_k / \tau)}{\sum_{\substack{j \in A(i)}} \dfrac{1}{\alpha_{y_j}^{y_i}} exp(h_i \cdot h_j / \tau)} 
\end{aligned}
\end{equation}
where $A(i)=\left\{1,...,i-1\right\}$, the index of samples before $x_i$, $S(i)=\left\{k|k \in A(i) \cap y_i \\ =y_k\right\}$, and $\tau$ is a temperature parameter. $\alpha_{y_j}^{y_i}$ is the correlation coefficient between label $y_i$ and label $y_j$, such that the distance between two representation vectors inversely proportional to their label correlation as illustrated in Figure \ref{fig:framework}. 
In this way, we can model the correlations among different emotions softly.
Finally, we train the prompt tuning and soft contrastive learning jointly by optimizing the loss,
$
\mathcal{L}_{\mathrm{SoftCL}} = \frac{1}{n}\sum_{i=1}^n (\mathcal{L}_{p,i} + \mathcal{L}_{\mathrm{SoftCL}, i}
)$.

\begin{table*}[t!]
\centering
\small
\caption{The main results of few-shot sentiment analysis.}
\vspace{-2mm}
\label{table:main results}
\setlength{\tabcolsep}{2.7mm}{\begin{tabular}{l|cc|cc|cc|cc|cc}
\hlineB{4}
                & \multicolumn{2}{c|}{K=1} & \multicolumn{2}{c|}{K=5} & \multicolumn{2}{c|}{K=10} & \multicolumn{2}{c|}{K=15} & \multicolumn{2}{c}{K=20} \\ 
           & Acc         & F1          & Acc         & F1          & Acc         & F1          & Acc         & F1          & Acc         & F1          \\ \hline
Supervised       & 0.145       & 0.090       & 0.320       & 0.255       & 0.410       & 0.330       & 0.457       & 0.380       & 0.493       & 0.432       \\ 
PET    & 0.335           & 0.284           & 0.439           & 0.372          & 0.472           & 0.407           & 0.493           & 0.419           & 0.515           & 0.445           \\ 
P-tuning(mlp)    & 0.285       & 0.230       & 0.437       & 0.372       & 0.473       & 0.396       & 0.495       & 0.418       & 0.507       & 0.434       \\ 
P-tuning(lstm)   & 0.301      & 0.250       & 0.441       & 0.372       & 0.467       & 0.383       & 0.489       & 0.412       & 0.510           & 0.433           \\ 
iPET       & 0.333       & 0.282       & 0.445       & 0.379       & 0.475       & 0.409       & 0.496       & 0.422       & 0.512       & 0.442      \\ 
Basic Prompt  & 0.331       & 0.332       & 0.427       & 0.464       & 0.474       & 0.492       & 0.491       & 0.501       & 0.490       & 0.503\\
Mixed Template   & 0.324           & 0.335           & 0.407           &0.457           & 0.467           & 0.506           & 0.501           & 0.530           & 0.509           & 0.540           \\ 
Soft Verbalizers & 0.093           & 0.084           & 0.161           & 0.185           & 0.275           & 0.306           & 0.354           & 0.360           & 0.449           & 0.479           \\ 
Soft Template & 0.273           & 0.248           & 0.404           & 0.425           & 0.416           & 0.451           & 0.508           & 0.523           & 0.512           & 0.557           \\ 
ChatGPT & 0.322           & 0.355           & 0.323           & 0.362          & -           & -           & -          & -           & -           & -           \\ 
\hline
\texttt{SCP}    & \textbf{0.347}       & \textbf{0.351}       & \textbf{0.446}       & \textbf{0.472}       & \textbf{0.480}       & \textbf{0.512}       & \textbf{0.519}      & \textbf{0.537}       & \textbf{0.529}       & \textbf{0.560}       \\ 
\texttt{SCP} (w/o SCoT) & 0.335       & 0.340       & 0.398       & 0.462       & 0.440       & 0.463       & 0.495       & 0.519       & 0.501       & 0.526\\
\texttt{SCP} (w/o SoftCL) & 0.342       & 0.338       & 0.431       & 0.464       & 0.465       & 0.498       & 0.480       & 0.499       & 0.526       & 0.555\\
\texttt{SCP} (w/o CL) & 0.331       & 0.344       & 0.420       & 0.450       & 0.461       & 0.500       & 0.467       & 0.503       & 0.507       & 0.552\\
\hlineB{4}
\end{tabular}}
\vspace{-2mm}
\end{table*}

\vspace{-2mm}
\section{Experiments}
\vspace{-2mm}
\subsection{Experimental Setups}
\vspace{-1mm}
\textbf{Datasets and Metrics.}
For lack of standard few-shot sentiment analysis datasets, we build few-shot sentiment analysis datasets based on a fine-grained emotion dataset GoEmotions, which contains 58K Reddit comments with 27 emotions and neutral \cite{Demszky2020GoEmotionsAD}. 
Specifically, we randomly select $K$ samples for each class from the original training set for training, 500 samples from the original training set for validating, and 5\% samples from the original test set for testing.
We set $K$ as 1, 5, 10, 15, and 20 in our experiments. 
Following the \cite{schick2021exploiting}, we adopt two popular metrics, accuracy (Acc) and macro-F1 (F1) to evaluate the performance.


\textbf{Baselines.}
To verify the effectiveness of our \texttt{SCP}, we compare it with one general supervised model and eight state-of-the-art few-shot text classification baselines, which are listed as follows:
Supervised \cite{devlin-etal-2019-bert}, iPet \cite{schick2021exploiting}, P-tuning(mlp) \cite{Liu2021GPTUT}, P-tuning(lstm) \cite{Liu2021GPTUT}, Basic Prompt \cite{ding2021openprompt}, Mixed Template \cite{Gu2021PPTPP}, Soft Template \cite{Lester2021ThePO}, Soft Verbalizers \cite{ding2021openprompt}, ChatGPT\footnote{For limitation of sequence length, we do the experiments on ChatGPT with one-shot and five-shot setting.} \cite{schulman2022chatgpt}. 

\textbf{Implementation Details.}
We use BERT base \cite{li2020bert} as our basic language model. 
For all experiments, the model is trained with a learning rate of $5 \times 10^{-5}$. The batch size is 4, the maximum sequence length is 256 and the temperature is 2.0. 

\vspace{-2mm}
\subsection{Experimental Results} 
\vspace{-1mm}
\textbf{Main Results.} Table \ref{table:main results} shows the results for our model and the compared methods.
From this table, we obtain the following findings.
1) The baseline Supervised method, which finetunes using general supervised classification, performs worse than prompt-based models (e.g., iPET, P-tuning, Basic Prompt), especially when the $K$ is small. Prompt-based model can help to capture the knowledge in pre-trained models.
2) Our \texttt{SCP} model outperforms all the strong few-shot text classification baselines regarding accuracy and F1 over all datasets.
This observation shows that our SCoT can enhance the model to infer the sentiment with multiple reasoning steps. Also, we model the correlations among different sentiment classes via SoftCL. 
3) Additionally, few-shot sentiment analysis is still a challenge for recent large language models, such as ChatGPT.


\textbf{Ablation Studies.} To demonstrate the effectiveness of the main components of \texttt{SCP}, we remove SCoT (\texttt{SCP} w/o SCoT) and SoftCL (\texttt{SCP} w/o SoftCL) from \texttt{SCP} (Table \ref{table:main results}). 
Removing SCoT will reduce the performance of \texttt{SCP}, which indicates SCoT helps predict the sentiment precisely by inferring it into multiple steps. 
Also, we can find that integrating soft contrastive learning into our model can improve accuracy and F1 score effectively.
It indicates that soft contrastive learning helps to model the relationships among the labels, which improves the representations of the sentences.

\begin{figure}[htp]
\vspace{-3mm}
    \centering
    \subfigure[\texttt{SCP}]{
    \label{subfig:SCP}
    \includegraphics[width=3.6cm,height=3.6cm]{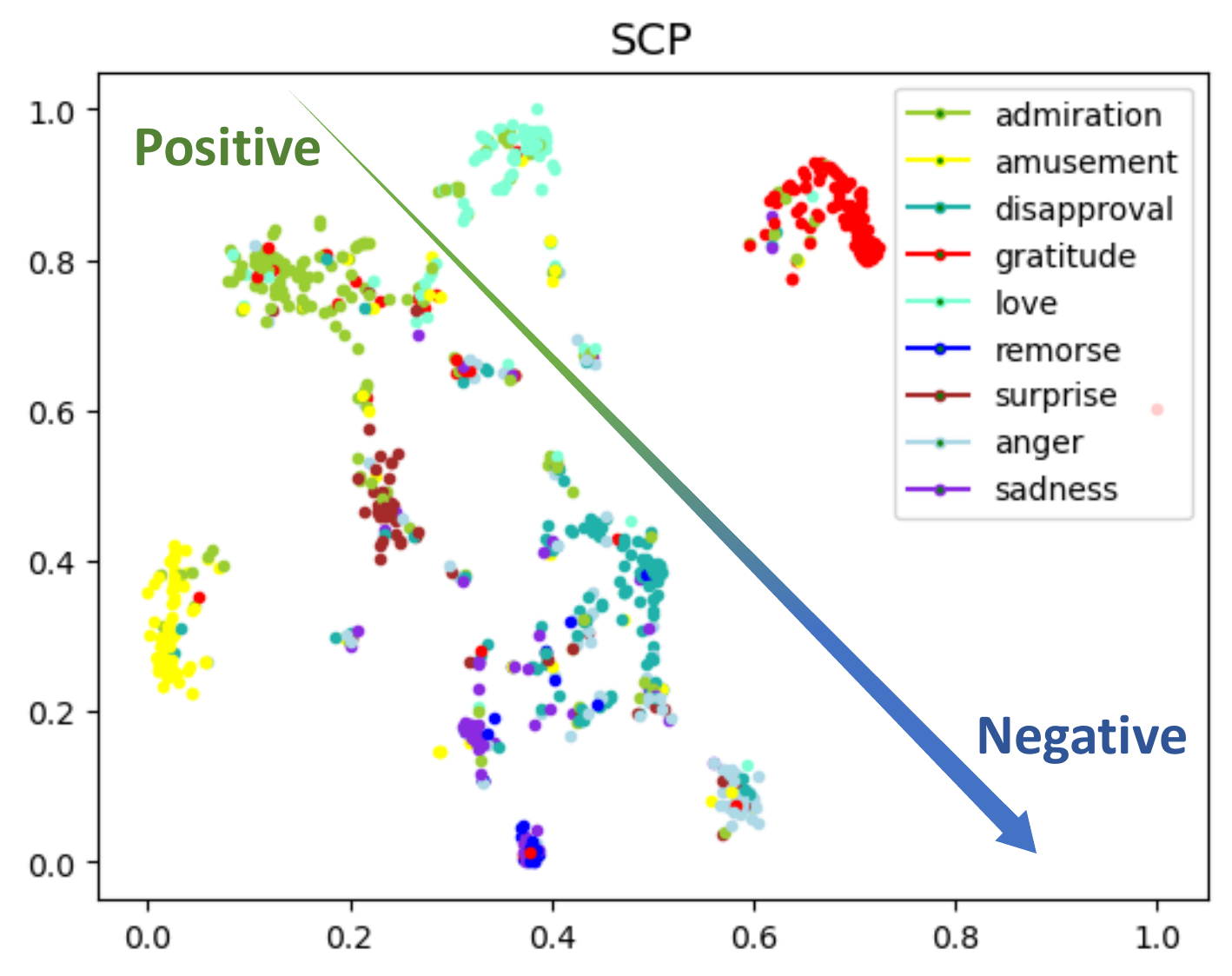}
    }
    \subfigure[\texttt{SCP} (w/o SoftCL)]{
    \label{BERT}
    \includegraphics[width=3.6cm,height=3.6cm]{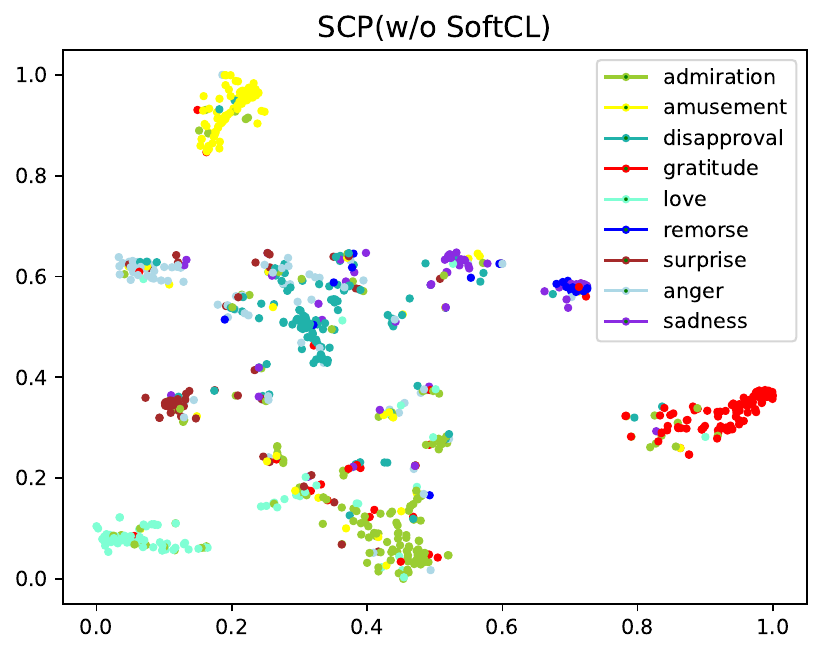}
    }
    \vspace{-3mm}
    \caption{The visualization of \texttt{SCP}. We use t-SNE to translate the ``[CLS]" representation to a two-dimension vector.}
    \label{fig:visualization}
    \vspace{-5mm}
\end{figure}

\textbf{Visualization.} We visualize the sentence representations of \texttt{SCP} and \texttt{SCP} w/o SoftCL over testing data in Figure \ref{fig:visualization}. 
We only visual nine classes to better display the figures. 
Our \texttt{SCP} model pushes representations of samples belonging to highly correlated labels (e.g., ``love", ``admiration") close and pulls representations of samples belonging to less correlated labels (e.g., ``love", ``anger") far. 
From the upper left to the lower right of Figure \ref{subfig:SCP}, there is an obvious distribution of emotions from positive to negative, from ``love" and ``admiration" and then ``disapproval" and ``anger". Ambiguous sentiments such as "surprise" are closer to positive than negative in the figure, which agrees with the findings in \cite{Demszky2020GoEmotionsAD}.

\vspace{-2mm}
\section{Related Work}
\vspace{-1mm}
\label{sect:related work}
\textbf{Sentiment Analysis.}
Sentiment classification (SC) aims to predict the sentiment label for the given sentence \cite{liu2012sentiment,zhou2020sentix,hu2018snnn}, which is a fundamental task of NLP. 
Lexicon-based approaches \cite{strapparava2004wordnet} and machine learning approaches \cite{strapparava2008learning} are used to infer the sentiment. 
The current SOTA methods make use of pre-trained language models (e.g., BERT \cite{devlin-etal-2019-bert}) for calculating input representations. 
Some recent studies try to apply large foundation models such as ChatGPT to sentiment analysis with zero or few samples given in the prompt \cite{Amin2023WillAC}. However, it also faces challenges in specific tasks \cite{Qin2023IsCA}. 

\textbf{Few-shot Text Classification.}
Few-shot learning can effectively reduce the number of labeled data for model training. 
A popular method in the field of few-shot text classification is to pre-train on a large corpus and then fine-tune on a specific task~\cite{Bao2020FewshotTC,Brown2020LanguageMA}. 
Some studies try prompt-based models to use knowledge in pre-trained models, which gives great improvements over standard supervised finetuning \cite{Schick2021ExploitingCF}. 
However, few studies focus on few-shot sentiment analysis, which differs from general few-shot text classification tasks.

\textbf{Contrastive Learning.}
Contrastive learning focuses on learning the features of instances between different instances by pulling samples with similar sentiments together and pushing samples with different sentiments apart \cite{He2020MomentumCF}. 
In NLP, self-supervised contrastive learning is considered to construct meaningful sentence representation effectively \cite{khosla2020supervised,Wang2021CLINECL}. 
Janine et al. \cite{Thoma2020SoftCL} propose the positiveness and negativeness scores to calculate soft contrastive loss with given anchors. Zhu et al. \cite{Zhu2022MultigranularityIC} combines multi-granularity classification loss with contrastive learning.
However, the semantic correlation between classes is not well studied in contrastive learning.

\textbf{Chain of Thought Learning.}
Several prior studies have shown that large language models can benefit from step-by-step reasoning by using a chain of thought prompting in zero-shot and few-shot settings \cite{Wei2022ChainOT,Kojima2022LargeLM}. With a suitable prompt, the language model is able to generate several intermediate steps, which significantly benefit the final result \cite{DBLP:conf/iclr/ZhouSHWS0SCBLC23}. 
Inspired by the prior work, we explored the chain of thought learning in sentiment analysis tasks.

\vspace{-2mm}
\section{Conclusions and Future Work}
\vspace{-1mm}
In this paper, we propose a Soft Contrastive learning-based Prompt (\texttt{SCP}) model to fully use the correlation of the sentiment labels. 
Particularly, we first enhance the model to infer sentiment step by step via a sentiment-aware CoT prompt module. Then, we calculate the distance between the positive and negative samples softly by considering the correlation of sentiment classes.
We compare our \texttt{SCP} model with several competitive methods.
The results 
show that our model can capture the relationships among the sentiment labels well.
In future work, we would like to extend our method to other fine-grained classification tasks.

\vspace{-2mm}
\section*{\normalsize{Acknowledgement}}
\vspace{-1mm}
This research is funded by the National Natural Science Foundation of China (No.62076069).


\vfill\pagebreak

\small
\bibliographystyle{IEEEbib}
\bibliography{refs}

\begin{thebibliography}{10}

\bibitem{liu2012sentiment}
Bing Liu,
\newblock ``Sentiment analysis and opinion mining,''
\newblock {\em Synthesis lectures on human language technologies}, vol. 5, no. 1, pp. 1--167, 2012.

\bibitem{chen2020mixtext}
Jiaao Chen, Zichao Yang, and Diyi Yang,
\newblock ``Mixtext: Linguistically-informed interpolation of hidden space for semi-supervised text classification,''
\newblock in {\em Proceedings of ACL}, 2020, pp. 2147--2157.

\bibitem{xie2020unsupervised}
Qizhe Xie, Zihang Dai, Eduard Hovy, Thang Luong, and Quoc Le,
\newblock ``Unsupervised data augmentation for consistency training,''
\newblock {\em Proceedings of NeurIPs}, vol. 33, pp. 6256--6268, 2020.

\bibitem{Schick2021ExploitingCF}
Timo Schick and Hinrich Sch{\"u}tze,
\newblock ``Exploiting cloze-questions for few-shot text classification and natural language inference,''
\newblock in {\em Proceedings of EACL}, 2021.

\bibitem{hu2021knowledgeable}
Shengding Hu, Ning Ding, Huadong Wang, Zhiyuan Liu, Jingang Wang, Juanzi Li, Wei Wu, and Maosong Sun,
\newblock ``Knowledgeable prompt-tuning: Incorporating knowledge into prompt verbalizer for text classification,''
\newblock in {\em Proceedings of ACL}, 2022, pp. 2225--2240.

\bibitem{Demszky2020GoEmotionsAD}
Dorottya Demszky, Dana Movshovitz-Attias, Jeongwoo Ko, Alan Cowen, Gaurav Nemade, and Sujith Ravi,
\newblock ``Goemotions: A dataset of fine-grained emotions,''
\newblock in {\em Proceedings of ACL}, 2020, pp. 4040--4054.

\bibitem{DBLP:conf/iclr/ZhouSHWS0SCBLC23}
Denny Zhou, Nathanael Sch{\"{a}}rli, Le~Hou, Jason Wei, Nathan Scales, Xuezhi Wang, Dale Schuurmans, Claire Cui, Olivier Bousquet, Quoc~V. Le, and Ed~H. Chi,
\newblock ``Least-to-most prompting enables complex reasoning in large language models,''
\newblock in {\em Proceedings of ICLR}, 2023.

\bibitem{khosla2020supervised}
Prannay Khosla, Piotr Teterwak, Chen Wang, Aaron Sarna, Yonglong Tian, Phillip Isola, Aaron Maschinot, Ce~Liu, and Dilip Krishnan,
\newblock ``Supervised contrastive learning,''
\newblock {\em Proceedings of NeurIPs}, vol. 33, pp. 18661--18673, 2020.

\bibitem{devlin-etal-2019-bert}
Jacob Devlin, Ming-Wei Chang, Kenton Lee, and Kristina Toutanova,
\newblock ``{BERT}: Pre-training of deep bidirectional transformers for language understanding,''
\newblock in {\em Proceedings of NAACL}, June 2019, pp. 4171--4186.

\bibitem{Wei2022ChainOT}
Jason Wei, Xuezhi Wang, Dale Schuurmans, Maarten Bosma, Ed~Chi, Quoc Le, and Denny Zhou,
\newblock ``Chain of thought prompting elicits reasoning in large language models,''
\newblock {\em ArXiv}, vol. abs/2201.11903, 2022.

\bibitem{parrott2001emotions}
W~Gerrod Parrott,
\newblock {\em Emotions in social psychology: Essential readings},
\newblock psychology press, 2001.

\bibitem{schick2021exploiting}
Timo Schick and Hinrich Sch{\"u}tze,
\newblock ``Exploiting cloze-questions for few-shot text classification and natural language inference,''
\newblock in {\em Proceedings of EACL}, 2021, pp. 255--269.

\bibitem{Liu2021GPTUT}
Xiao Liu, Yanan Zheng, Zhengxiao Du, Ming Ding, Yujie Qian, Zhilin Yang, and Jie Tang,
\newblock ``Gpt understands, too,''
\newblock {\em ArXiv}, vol. abs/2103.10385, 2021.

\bibitem{ding2021openprompt}
Ning Ding, Shengding Hu, Weilin Zhao, Yulin Chen, Zhiyuan Liu, Haitao Zheng, and Maosong Sun,
\newblock ``Openprompt: An open-source framework for prompt-learning,''
\newblock in {\em Proceedings of ACL}, 2022, pp. 105--113.

\bibitem{Gu2021PPTPP}
Yuxian Gu, Xu~Han, Zhiyuan Liu, and Minlie Huang,
\newblock ``Ppt: Pre-trained prompt tuning for few-shot learning,''
\newblock in {\em Proceedings of ACL}, 2022, pp. 8410--8423.

\bibitem{Lester2021ThePO}
Brian Lester, Rami Al-Rfou, and Noah Constant,
\newblock ``The power of scale for parameter-efficient prompt tuning,''
\newblock in {\em Proceedings of EMNLP}, 2021, pp. 3045--3059.

\bibitem{schulman2022chatgpt}
John Schulman, B~Zoph, C~Kim, J~Hilton, J~Menick, J~Weng, JFC Uribe, L~Fedus, L~Metz, M~Pokorny, et~al.,
\newblock ``Chat{GPT}: Optimizing language models for dialogue,''
\newblock in {\em OpenAI blog}, 2022.

\bibitem{li2020bert}
Linyang Li, Ruotian Ma, Qipeng Guo, Xiangyang Xue, and Xipeng Qiu,
\newblock ``Bert-attack: Adversarial attack against bert using bert,''
\newblock in {\em Proceedings of EMNLP}, 2020, pp. 6193--6202.

\bibitem{zhou2020sentix}
Jie Zhou, Junfeng Tian, Rui Wang, Yuanbin Wu, Wenming Xiao, and Liang He,
\newblock ``Sentix: A sentiment-aware pre-trained model for cross-domain sentiment analysis,''
\newblock in {\em Proceedings of COLING}, 2020, pp. 568--579.

\bibitem{hu2018snnn}
Qinmin Hu, Jie Zhou, Qin Chen, and Liang He,
\newblock ``Snnn: promoting word sentiment and negation in neural sentiment classification,''
\newblock in {\em Proceedings of AAAI}, 2018, vol.~32.

\bibitem{strapparava2004wordnet}
Carlo Strapparava, Alessandro Valitutti, et~al.,
\newblock ``Wordnet affect: an affective extension of wordnet.,''
\newblock in {\em Proceedings of Lrec}. Citeseer, 2004, vol.~4, p.~40.

\bibitem{strapparava2008learning}
Carlo Strapparava and Rada Mihalcea,
\newblock ``Learning to identify emotions in text,''
\newblock in {\em Proceedings of SAC}, 2008, pp. 1556--1560.

\bibitem{Amin2023WillAC}
Mostafa~M. Amin, E.~Cambria, and Bj{\"o}rn Schuller,
\newblock ``Will affective computing emerge from foundation models and general ai? a first evaluation on chatgpt,''
\newblock {\em ArXiv}, vol. abs/2303.03186, 2023.

\bibitem{Qin2023IsCA}
Chengwei Qin, Aston Zhang, Zhuosheng Zhang, Jiaao Chen, Michihiro Yasunaga, and Diyi Yang,
\newblock ``Is chatgpt a general-purpose natural language processing task solver?,''
\newblock {\em ArXiv}, vol. abs/2302.06476, 2023.

\bibitem{Bao2020FewshotTC}
Yujia Bao, Menghua Wu, Shiyu Chang, and Regina Barzilay,
\newblock ``Few-shot text classification with distributional signatures,''
\newblock {\em ArXiv}, vol. abs/1908.06039, 2020.

\bibitem{Brown2020LanguageMA}
Tom Brown, Benjamin Mann, Nick Ryder, Melanie Subbiah, Jared~D Kaplan, Prafulla Dhariwal, Arvind Neelakantan, Pranav Shyam, Girish Sastry, Amanda Askell, et~al.,
\newblock ``Language models are few-shot learners,''
\newblock {\em Proceedings of NeurIPs}, vol. 33, pp. 1877--1901, 2020.

\bibitem{He2020MomentumCF}
Kaiming He, Haoqi Fan, Yuxin Wu, Saining Xie, and Ross~B. Girshick,
\newblock ``Momentum contrast for unsupervised visual representation learning,''
\newblock {\em Proceedings of CVPR}, pp. 9726--9735, 2020.

\bibitem{Wang2021CLINECL}
Dong Wang, Ning Ding, Piji Li, and Haitao Zheng,
\newblock ``Cline: Contrastive learning with semantic negative examples for natural language understanding,''
\newblock in {\em Proceedings of ACL}, 2021, pp. 2332--2342.

\bibitem{Thoma2020SoftCL}
Janine Thoma, Danda~Pani Paudel, and Luc~Van Gool,
\newblock ``Soft contrastive learning for visual localization,''
\newblock in {\em Neural Information Processing Systems}, 2020.

\bibitem{Zhu2022MultigranularityIC}
Peng~Fei Zhu, Wanying Zhang, Yu~Wang, and Qinghua Hu,
\newblock ``Multi-granularity inter-class correlation based contrastive learning for open set recognition,''
\newblock {\em Int. J. Softw. Informatics}, vol. 12, pp. 157--175, 2022.

\bibitem{Kojima2022LargeLM}
Takeshi Kojima, Shixiang~Shane Gu, Machel Reid, Yutaka Matsuo, and Yusuke Iwasawa,
\newblock ``Large language models are zero-shot reasoners,''
\newblock {\em ArXiv}, vol. abs/2205.11916, 2022.

\end{thebibliography}

\end{document}